\def\year{2015}
\documentclass[letterpaper]{article}
\usepackage{aaai}
\usepackage{times}
\usepackage{helvet}
\usepackage{courier}
\usepackage{amsmath}
\usepackage{amssymb}
\usepackage{makecell}

\begin{document}
%
\title{Semisupervised Autoencoder for Sentiment Analysis}
 \author{Shuangfei Zhai \;  Zhongfei (Mark) Zhang\\
Computer Science Department, Binghamton University\\
4400 Vestal Pkwy E, Binghamton, NY 13902\\
szhai2@binghamton.edu \; zhongfei@cs.binghamton.edu\\
 }
\maketitle
\begin{abstract}
\begin{quote}
In this paper, we investigate the usage of autoencoders in modeling textual data. Traditional autoencoders suffer from at least two aspects: scalability with the high dimensionality of vocabulary size and dealing with task-irrelevant words. We address this problem by introducing supervision via the loss function of autoencoders. In particular, we first train a linear classifier on the labeled data, then define a loss for the autoencoder with the weights learned from the linear classifier. To reduce the bias brought by one single classifier, we define a posterior probability distribution on the weights of the classifier, and derive the marginalized loss of the autoencoder with Laplace approximation. We show that our choice of loss function can be rationalized from the perspective of Bregman Divergence, which justifies the soundness of our model. We evaluate the effectiveness of our model on six sentiment analysis datasets, and show that our model significantly outperforms all the competing methods with respect to classification accuracy. We also show that our model is able to take advantage of unlabeled dataset and get improved performance. We further show that our model successfully learns highly discriminative feature maps, which explains its superior performance.
\end{quote}
\end{abstract}

\section{Introduction} \label{introduction}

In machine learning, documents are usually represented as Bag of Words (BoW), which nicely reduces a piece of text with arbitrary length to a fixed length vector. Despite its simplicity, BoW remains the dominant representation in many applications including text classification. There has also been a large body of work dedicated to learning useful representations for textual data \cite{vsm,lda,lsa,word2vec,xavier}. By exploiting the co-occurrence pattern of words, one can learn a low dimensional vector that forms a compact and meaningful representation for a document. The new representation is often found useful for subsequent tasks such as topic visualization and information retrieval. In this paper, we investigate the application of one of the most popular representation learning methods, namely autoencoders \cite{bengio}, to learn task-dependent representations for textual data. Our model differs from most of the existing work as it naturally incorporates label information into its objective function, which allow the learned representation to be directly coupled with the task of interest.

In this paper we focus on a specific class of task in text mining: Sentiment Analysis (SA). We further focus on a special case of SA as a binary classification problem, where a given piece of text is either of positive or negative attitude. This problem is interesting largely due to the emergence of online social networks, where people consistently express their opinions about certain subjects. Also, it is easy to obtain a large amount of clean labeled data for SA by crawling reviews from websites such as IMDB or Amazon. Thus, SA is an ideal benchmark for evaluating text classification models (and features). 

Autoencoders have attracted a lot of attention in recent years as a building block of Deep Learning \cite{bengio}. They act as the feature learning methods by reconstructing inputs with respect to a given loss function. In a neural network implementation of autoencoders, the hidden layer is taken as the learned feature. While it is often trivial to obtain good reconstructions with plain autoencoders, much effort has been devoted on regularizations in order to prevent them against overfitting \cite{bengio,dae,cae}. However, little attention has been devoted to the loss function, which we argue is critical for modeling textual data. The problem with the commonly adopted loss functions (squared Euclidean distance and element-wise KL Divergence, for instance) is that they try to reconstruct all dimensions of input independently and undiscriminatively. However, we argue that this is not the optimal approach when our interest is text classification. The reason is two folds. First, it is well known that in natural language the distribution of word occurrences follows the power-law. This means that a few of the most frequent words will account for most of the probability mass of word occurrences. An immediate result is that the Autoencoder puts most of its effort on reconstructing the most frequent words well but (to a certain extent) ignores the less frequent ones. This may lead to a bad performance especially when the class distribution is not well captured by merely the frequent words. For sentiment analysis, this problem is especially severe because it is obvious that the truly useful features (words or phrases expressing a clear polarity) only occupy a small fraction of the whole vocabulary; and reconstructing irrelevant words such as 'actor' or 'movie' very well is not likely to help learn more useful representations to classify the sentiment of movie reviews. Second, explicitly reconstructing all the words in an input text is expensive, because the latent representation has to contain all aspects of the semantic space carried by the words, even if they are completely irrelevant. As the vocabulary size can easily reach the range of tens of thousands even for a moderate sized dataset, the hidden layer size has to be chosen very large to obtain a reasonable reconstruction, which causes a huge waste of model capacity and makes it difficult to scale to large problems.

In fact, the reasoning above applies to all the unsupervised learning methods in general, which we argue is one of the most important problems to address in order to learn task-specific representations. This naturally leads us to the semisupervised approach, where label information is introduced to guide the feature learning procedure. In particular, we propose a novel loss function for training autoencoders that are directly coupled with the classification task. We first train a linear classifier on BoW, then a Bregman Divergence \cite{bregman} is derived as the loss function of a subsequent autoencoder. The new loss function gives the autoencoder the information about directions along which the reconstruction should be accurate, and where larger reconstruction errors are tolerated. Informally, this can be considered as a weighting of words based on their correlations with the class label: predictive words should be given large weights in the reconstruction even they are not frequent words, and vice versa. Furthermore, to reduce the bias introduced by the linear classifier, we take a Bayesian view by defining a posterior distribution on the weights of the classifier. We then approximate the posterior with Laplace approximation and derive the marginalized loss function for the autoencoder. We show that our model successfully learns features that are highly discriminative with respect to class labels, and also outperform all the competing methods evaluated by classification accuracy. Moreover, the derived loss can also be applied to unlabeled data, which allows the model to learn further better representations.

\section{Model} \label{model}
\subsection{Denoising Autoencoders} \label{intro:ae}

Autoencoders learn functions that can reconstruct the inputs. They are typically implemented as a neural network with one hidden layer, and one can extract the activation of the hidden layer as the new representation. Mathematically, we are given a collection of data points $X = \{x_i\}, x_i \in R^d, i \in [1, m]$, the objective function of an autoencoder is thus:
\begin{equation} \label{eq:ae}
\begin{split}
\min &\sum_i D(\tilde{x}_i, x_i) \\
s.t. \quad &h_i = g(Wx_i + b), \tilde{x}_i = f(W'h_i + b')
\end{split}
\end{equation}
where $W \in R^{k \times d}, b \in R^{k}, W' \in R^{d \times k}, b' \in R^{d}$ are the parameters to be learned; $D$ is a loss function, such as the squared Euclidean Distance $\|\tilde{x} - x\|_2^2$; $g$ and $f$ are predefined nonlinear functions, which we set as $g(x) = \max(0, x)$, $f(x) = (1 + exp(-x))^{-1}$ in this paper; $h_i$ is the learned representation; $\tilde{x}_i$ is the reconstruction. A common approach is to use tied weights by setting $W = W'$; this usually works better as it speeds up learning and prevents overfitting at the same time. For this reason, we always use tied weights in this paper.

Autoencoders transform an unsupervised learning problem to a supervised one by the self reconstruction criteria. This enables one to use all the tools developed for supervised learning such as back propagation to efficiently train the autoencoders. Moreover, thanks to the nonlinear functions $f$ and $g$, autoencoders are able to learn non-linear and possibly overcomplete representations, which give the model much more expressive power than their linear counter parts such as PCA (LSA) \cite{lsa}.

In this paper, we adopt one of the most popular variants of autoencoders, namely Denoising Autoencoder. Denoising Autoencoder works by reconstructing the input from a noised version of itself. The intuition is that a robust model should be able to reconstruct the input well even in the presence of noises, due to the high correlation among features. For example, imagine deleting or adding a few words from/to a document, the semantics should still remain unchanged, thus the autoencoder should learn a consistent representation from all the noisy inputs. In the high level, Denoising Autoencoders are equivalent to ordinary autoencoders trained with dropout \cite{dropout}, which has been shown as an effective regularizer for (deep) neural networks. Formally, let $q(\bar{x} | x)$ be a predefined noising distribution, and $\bar{x}$ be a noised sample of $x$: $\bar{x} \sim q(\bar{x} | x)$. The objective function takes the form of sum of expectations over all the noisy samples:
\begin{equation}\label{eq:daeloss}
\begin{split}
\min &\sum_i \mathrm{E}_{q(\bar{x}_i | x_i)}D(\tilde{x}_i, x_i) \\
s.t. \quad & h_i = g(W\bar{x}_i + b), \tilde{x}_i = f(W'h_i + b')
\end{split}
\end{equation}
where we have slightly overloaded the notation to let $\tilde{x}_i$ denote the reconstruction calculated from the noised input $\bar{x}_i$. While the marginal objective function requires infinite many noised samples per data point, in practice it is sufficient to simulate it stochastically. That is, for each example seen in the stochastic gradient descent training, we randomly sample a $\bar{x}_i$ from $q(\bar{x}_i | x_i)$ and calculate the gradient with ordinary back propagation.


\subsection{Loss Function as Bregman Divergence} \label{intro:bregman}

We then discuss the proper choice of the loss function $D$ in \eqref{eq:daeloss} as a specific form of Bregman Divergence. Bregman Divergence \cite{bregman} generalizes the notion of distance in a $d$ dimensional space. To be concrete, given two data points $\tilde{x}, x \in R^d$ and a convex function $f(x)$ defined on $R^d$, the Bregman Divergence of $\tilde{x}$ from $x$ with respect to $f$ is: 
\begin{equation}
D_f(\tilde{x}, x) = f(\tilde{x}) - (f(x) + \nabla{f(x)}^T(\tilde{x} - x)).
\end{equation}
Namely, Bregman Divergence measures the distance between two points $\tilde{x}, x$ as the deviation between the function value of $f$ and the linear approximation of $f$ around $x$ at $\tilde{x}$. 

Two of the most commonly used loss functions for autoencoders are the squared Euclidean distance and element-wise KL divergence. It is not difficult to verify that they both fall into this family by choosing $f$ as the squared $\ell_2$ norm and the sum of element-wise entropy respectively. What the two loss functions have in common is that they make no distinction among dimensions of the input. In other words, each dimension of the input is pushed to be reconstructed equally well. While autoencoders trained in this way have been shown to work very well on image data, learning much more interesting and useful features than the original pixel intensity features, they are less appropriate for modeling textual data. The reason is two folds. First, textual data are extremely sparse and high dimensional, where the dimensionality is equal to the vocabulary size. To maintain all the information of the input in the hidden layer, a very large layer size must be adopted, which makes the training cost extremely large. Second, ordinary autoencoders are not able to deal with the power law of word distributions, where a few of the most frequent words account for most of the word occurrences. As a result, frequent words naturally gain favor to being reconstructed accurately, and rare words tend to be reconstructed with less precision. This problem is also analogous to the imbalanced classification setting. This is especially problematic when frequent words carry little information about the task of interest, which is not uncommon. Examples include stop words (\textit{the, a, this, from}) and topic related terms (\textit{movie, watch, actress}) in a movie review sentiment analysis task. 
 
\subsection{Semisupervised Autoencoder with Bregman Divergence} \label{model}

To address the problems mentioned above, we propose to introduce supervision to the training of autoencoders. To achieve this, we first train a linear classifier on Bag of Words, and then use the weight of the learned classifier to define a new loss function for the autoencoder. Now let us first describe our choice of loss function, and then elaborate the motivation later:
\begin{equation} \label{eq:df}
D(\tilde{x}, x) = (\theta^T(\tilde{x} - x))^2.
\end{equation}
where $\theta \in R^d$ are the weights of the linear classifier, and we have omitted the bias for simplicity. Before we delve into more details, note that Equation \eqref{eq:df} is a valid distance, as it is non-negative and reaches zeros if and only if $\tilde{x} = x$. Moreover, the reconstruction error is only measured after projecting on $\theta$; this guides the reconstruction to be accurate only along directions where the linear classifier is sensitive to. Note also that  Equation \eqref{eq:df} on the one hand uses label information ($\theta$ has been trained with labeled data), on the other hand no explicit labels are directly referred to (only requires $x_i$). Thus one is able to train an autoencoder on both labeled and unlabeled data with the loss function in Equation \eqref{eq:df}. This subtlety distinguishes our method from pure supervised or unsupervised learning, and allows us to enjoy the benefit from both worlds. 

As a design choice, we consider SVM with squared hinge loss (SVM2) and $\ell_2$ regularization as the linear classifier, but other classifiers such as Logistic Regression can be used and analyzed similarly. Let us denote $\{x_i\}, x_i \in R^d$ as the collection of samples, and $\{y_i\}, y_i \in \{1, -1\}$ as the class labels; the objective function SVM2 is:
\begin{equation} \label{eq:svm}
L(\theta) = \sum_i (\max(0, 1 - y_i \theta^T x_i))^2 + \lambda \|\theta\|^2.
\end{equation}
Here $\theta \in R^d$ is the weight; $\lambda$ is the weight decay parameter.

Equation \eqref{eq:svm} is continuous and differentiable everywhere with respect to $\theta$, so the model can be easily trained with stochastic gradient descent. The next (and most critical) step of our approach is to transfer label information from the linear classifier to the autoencoder. With this in mind, we examine the loss induced by each sample as a function of the input, while with $\theta$ fixed:
\begin{equation}
f(x_i) = (\max(0, 1 - y_i \theta^T x_i))^2
\end{equation}
Note that $f(x_i)$ is defined on the input space $R^d$, which should be contrasted with $L(\theta)$ in Equation \eqref{eq:svm} which is a function of $\theta$. We are interested in $f(x_i)$ because if we consider moving each input $x_i$ to $\tilde{x}_i$, $f(x_i)$ indicates the direction along which the loss is sensitive to. If we think of $\tilde{x}$ as the reconstruction of $x_i$ obtained from an autoencoder, a good $\tilde{x}_i$ should be in a way such that the deviation of $\tilde{x}_i$ from $x_i$ is small evaluated by $f(x_i)$. In other words, we would like $\tilde{x}_i$ to still be correctly classified by the pretrained linear classifier. Therefore, $f(x_i)$ should be a much better function to evaluate the deviation of two samples. if we can derive a Bregman Divergence from $f(x_i)$ and use it as the loss function of the subsequent autoencoder training, the autoencoder should be guided to give reconstruction errors that do not confuse the classifier. Note that $f(x_i)$ is a quadratic function of $x_i$ whenever $f(x_i) > 0$, so we only need to derive the Hessian matrix in order to achieve the Bregman Divergence. The Hessian follows as:

\begin{equation} \label{eq:hessian}
H(x_i) = 
    \begin{cases}
      \theta \theta^T, & \text{if}\ 1 - y_i\theta^Tx_i > 0 \\
      0, & \text{otherwise}.
    \end{cases}
\end{equation}
Recall that for a quadratic function with Hessian matrix $H$, the Bregman Divergence is simply $ (\tilde{x} - x)^TH(\tilde{x} - x)$; then we have:
\begin{equation} \label{eq:df0}
D(\tilde{x}_i, x_i) = 
    \begin{cases}
      (\theta^T(\tilde{x}_i - x_i))^2, & \text{if}\ 1 - y_i\theta^Tx_i > 0 \\
      0, & \text{otherwise}
    \end{cases}
\end{equation}
In words, Equation \eqref{eq:df0} says that we measure the reconstruction loss for difficult examples (those that satisfy $1 - y_i\theta^Tx_i > 0$) with Equation \eqref{eq:df}; and there is no reconstruction loss at all for easy examples. This discrimination is undesirable, because in this case the Autoencoder would completely ignore easy examples, and there is no way to guarantee that the $\tilde{x}_i$ can be correctly classified. Actually, this split is just an artifact of the hinge loss and the asymmetrical property of Bregman Divergence. Hence, we perform a simple correction by ignoring the condition in Equation \eqref{eq:df0}, which basically pretends that all the examples induce a loss. This directly yields the loss function as in Equation \eqref{eq:df}.

\subsection{The Bayesian Marginalization}
In principle, one may directly apply Equation \eqref{eq:df} as the loss function in place of the squared Euclidean distance and train an autoencoder. However, doing so might introduce a bias brought by one single classifier. As a remedy, we resort to the Bayesian approach, which defines a probability distribution over $\theta$. Although SVM2 is not a probabilistic classifier like Logistic Regression, we can borrow the idea of Energy Based Model \cite{bengio} and use $L(\theta)$ as the negative log likelihood of the following distribution:

\begin{equation}
p(\theta) = \frac{\exp(-\beta L(\theta))}{\int \exp(-\beta L(\theta)) d\theta}
\end{equation}
where $\beta > 0$ is the temperature parameter which controls the shape of the distribution $p$. Note that the larger $\beta$ is, the sharper $p$ will be. In the extreme case, $p(\theta)$ is reduced to a uniform distribution as $\beta$ approaches $0$, and collapses into a single $\delta$ function as $\beta$ goes to positive infinity.

Given $p(\theta)$, we rewrite Equation \eqref{eq:df} as an expectation over $\theta$:

\begin{equation} \label{eq:int}
\begin{split}
D(\tilde{x}, x) = \mathrm{E}_{\theta \sim p(\theta)} (\theta ^T(\tilde{x} - x))^2 
= \int (\theta^T(\tilde{x} - x))^2 p(\theta) d\theta.
\end{split}
\end{equation}
Obviously there is now no closed form expression for $D(\tilde{x}, x)$. To solve it one could use sampling methods such as MCMC, which provides unbiased estimates of the expectation but could be slow in practice. Instead, we use the Laplace approximation, which approximates $p(\theta)$ by a Gaussian distribution $\tilde{p}(\theta) = \mathcal{N}(\hat{\theta}, \Sigma)$. As estimating the full covariance matrix is prohibitive, we further constrain $\Sigma$ to be diagonal. The benefit of doing so is that the expectation can now be computed directly in closed form. To see this, by simply replacing $p(\theta)$ with $\tilde{p}(\theta)$ in Equation \eqref{eq:int}:
\begin{equation} \label{eq:int}
\begin{split}
D(\tilde{x}, x) = &\mathrm{E}_{\theta \sim \tilde{p}(\theta)} (\theta^T(\tilde{x} - x))^2\\ 
= &(\tilde{x} - x)^T \mathrm{E}_{\theta \sim \tilde{p}(\theta)}(\theta \theta^T) (\tilde{x} - x) \\
= &(\tilde{x} - x)^T (\hat{\theta} \hat{\theta}^T + \Sigma) (\tilde{x} - x) \\
= &(\hat{\theta}^T(\tilde{x} - x))^2 + ({\Sigma}^{\frac{1}{2}}(\tilde{x} - x))^T({\Sigma}^{\frac{1}{2}}(\tilde{x} - x)).
\end{split}
\end{equation}
where $D$ now involves two parts, corresponding to the mean and variance term of the Gaussian distribution respectively. Now let us derive $\tilde{p}(\theta)$ for $p(\theta)$. In Laplace approximation, $\hat{\theta}$ is chosen as the mode of $p(\theta)$, which is exactly the solution to the SVM2 optimization problem. For $\Sigma$, we have:
\begin{equation} \label{eq:df1}
\begin{split}
\Sigma = &(diag(\frac{\partial^2{L(\theta)}}{\partial{\theta^2}}))^{-1} \\
= &\frac{1}{\beta}(diag(\sum_i{\mathbb{I}(1 - y_i \theta^Tx_i > 0)} x_i^2))^{-1}
\end{split}
\end{equation}
Here we have overridden $diag$ but letting it denote a diagonal matrix induced either by a square matrix or a vector; $\mathbb{I}$ is the indicator function; $(\cdot)^{-1}$ denotes matrix inverse. Interestingly, the second term in Equation \eqref{eq:int} is now equivalent to the squared Euclidean distance after performing element-wise normalizing the input using all difficult examples. The effect of this normalization is that the reconstruction errors of frequent words are down weighted; on the other hand, discriminative words are given higher weights as they would occur less frequently in difficult examples. Note that it is important to use a relatively large $\beta$ in order to avoid the variance term dominating the mean term. In other words, we need to ensure $p(\theta)$ to be reasonable peaked around $\hat{\theta}$ to effective take advantage of label information.


\section{Experiments} \label{experiment}

\subsection{Datasets}
We evaluate our model on six Sentiment Analysis benchmarks. The first one is the IMDB dataset \footnote{http://ai.stanford.edu/~amaas/data/sentiment/} \cite{mass}, which consists of movie reviews collected from IMDB. The IMDB dataset is one of the largest sentiment analysis dataset that is publicly available; it also comes with an unlabeled set which allows us to evaluate semisupervised learning methods. The rest five datasets are all collected from Amazon \footnote{http://www.cs.jhu.edu/~mdredze/datasets/sentiment/}\cite{amazon}, which corresponds to the reviews of five different products: books, DVDs, music, electronics, kitchenware. All the six datasets are already tokenized as either uni-gram or bi-gram features. For computational reasons, we only select the words that occur in at least $30$ training examples. We summarize the statistics of datasets in Table \ref{tb:datasets}.

\begin{table}[t]
\scriptsize
\caption{Statistics of the datasets.}
\label{tb:datasets} 
\begin{center}
\begin{tabular}{lllllll}

& \bf IMDB & \bf books & \bf DVD & \bf music & \bf electronics  & \bf kitchenware\\ \hline
\# train & 25,000 & 10,000 & 10,000 & 18,000 & 6,000 & 6,000\\
\# test & 25,000 & 3,105 &  2,960 & 2,661 & 2,862 & 1,691\\
\# unlabeled & 50,000 & N/A &  N/A & N/A & N/A & N/A\\
\# features & 8,876 & 9,849 & 10,537 & 13,099 & 5,091 & 3,907\\
\% positive  & 50 & 49.81 & 49.85 & 50.16 & 49.78 & 50.08\\
\end{tabular}
\end{center}
\end{table}

\subsection{Methods}

\begin{itemize}
\item Bag of Words (BoW). Instead of using the raw word counts directly, we take a simple step of data normalization:
\begin{equation} \label{eq:normalization}
x_{i,j} = \frac{\log(1 + c_{i,j})}{\max_j \log(1 + c_{i,j})}
\end{equation}
where $c_{i, j}$ denotes the number of occurrences of the $j$\textit{th} word in the $i$\textit{th} document, $x_{i, j}$ denotes the normalized count. We choose this normalization because it preserves the sparsity of the Bag of Words features; also each feature element is normalized to the range $[0, 1]$. Note that the very same normalized Bag of Words features are fed into the autoencoders.

\item Denoising Autoencoder (DAE) \cite{dae}. This refers to the regular Denoising Autoencoder defined in Equation \eqref{eq:ae} with squared Euclidean distance loss: $D(\tilde{x}, x) = \|\tilde{x} - x\|_2^2$. This is also used in \cite{xavier} on the Amazon datasets for domain adaptation. We use ReLu $max(0, x)$ as the activation function, and Sigmoid as the decoding function.

\item Denoising Autoencoder with Finetuning (DAE+) \cite{dae}. This denotes the common approach to continue training an DAE on labeled data by replacing the decoding part of DAE with a Softmax layer. 

\item Feedforward Neural Network (NN). This is the standard fully connected neural network with one hidden layer and random initialization. We use the same activation function as that in Autoencoders, i.e., ReLU.

\item Logistic Regression with Dropout (LrDrop) \cite{lrdrop}. This is a model where logistic regression is regularized with the marginalized dropout noise. LrDrop differs from our approach as it uses feature noising as an explicit regularization. Another difference is that our model is able to learn nonlinear representations, not merely a classifier, and thus is potentially able to model more complicated patterns in data. 

\item Semisupervised Bregman Divergence Autoencoder (SBDAE). This corresponds to our model with Denoising Autoencoder as the feature learner. The training process is roughly equivalent to training on BoW followed by the training of DAE, except that the loss function of DAE is replaced with the loss function defined in Equation \eqref{eq:int}. We cross validate $\beta$ from the set $\{10^{4}, 10^{5}, 10^{6}, 10^{7}, 10^{8}\}$ (note that larger $\beta$ corresponds to weaker Bayesian regularization).

\item Semisupervised Bregman Divergence Autoencoder with Finetuning (SBDAE+).
\end{itemize}

Note that except for BoW and LrDrop, all the other methods require a predefined dimensionality of representation. We use fixed sizes on all the datasets. For SBDAE and NN, a small hidden size is sufficient, so we use $200$. For DAE, we observe that it benefits from very large hidden sizes; however, due to computational constraints, we take $2000$. For BoW, DAE, SBDAE, we use SVM2 as the classifier. All the models are trained with mini-batch Stochastic Gradient Descent with momentum of $0.9$.

\subsection{Results}

\begin{table*}[t]
\caption{Left: our model achieves the best results on four (large ones) out of six datasets. Right: our model is able to take advantage of unlabeled data and gain better performance.}
\label{tb:results} 
\begin{center}
\begin{tabular}{lllllll||l}
& \bf books & \bf DVD & \bf music & \bf electronics  & \bf kitchenware & \bf IMDB & \bf IMDB + unlabled \\ \hline
BoW &  10.76 & 11.82 & 11.80 & 10.41 & 9.34 & 11.48 &N/A\\
DAE &  15.10 & 15.64 & 15.44 & 14.74 & 12.48 & 14.60 &13.28\\
DAE+ &  11.40 & 12.09 & 11.80 & 11.53 & 9.23 & 11.48 &11.47\\
NN &  11.05 & 11.89 & 11.42 & 11.15 & 9.16 &11.60 & N/A\\
LrDrop &  9.53 & 10.95 & 10.90 & \bf 9.81 & \bf 8.69 &10.88 & 10.73\\
SBDAE &  \bf 9.16 & \bf 10.90 & \bf 10.59 & 10.02 & 8.87 &\bf 10.52 & \bf 10.42\\
SBDAE+ &  \bf 9.12 & \bf 10.90 & \bf 10.58 & 10.01 & 8.83 &\bf 10.50 & \bf 10.41\\

\end{tabular}
\end{center}
\end{table*}

We first summarize the results as in classification error rate in Table \ref{tb:results}. First of all, our model consistently beats BoW with a margin, and it achieves the best results on four (larger) datasets out of six. On the other hand, DAE, DAE+ and NN all fail to outperform BoW, although they share the same architecture as nonlinear classifiers. This suggests that SBDAE be able to learn a much better nonlinear feature transformation function by training with a more informed objective (than that of DAE). Moreover, note also that finetuning on labeled set (DAE+) significantly improves the performance of DAE, which is ultimately on a par with training a neural net with random initialization (NN). However, finetuning offers little help to SBDAE, as it is already implicitly guided by labels during the training.

LrDrop is the second best method that we have tested. Thanks to the usage of dropout regularization, it consistently outperforms BoW, and achieves the best results on two (smaller) datasets. Compared with LrDrop, it appears that our model works better on large datasets ($\approx 10 K$ words, more than $10K$ training examples) than smaller ones. This indicates that in high dimensional spaces with sufficient samples, SBDAE benefits from learning a nonlinear feature transformation that disentangles the underlying factors of variation, while LrDrop is incapable of doing so due to its nature as a linear classifier.

As the training of the autoencoder part of SBDAE does not require the availability of labels, we also try incorporating unlabeled data after learning the linear classifier in SBDAE. As shown in Table \ref{tb:results}, doing so further improves the performance over using labeled data only. This justifies that it is possible to bootstrap from a relatively small amount of labeled data and learn better representations with more unlabeled data with SBDAE. 

To gain more insights of the results, we further visualize the filters learned by SBDAE and DAE on the IMDB dataset in Table \ref{tb:topics}. In particular, we show the top $5$ most activated and deactivated words of the first $8$ filters (corresponding to the first $8$ rows of $W$) of SBDAE and DAE, respectively. First of all, it seems very difficult to make sense of the filters of DAE as they are mostly common words with no clear co-occurrence pattern. By comparison, if we look at the filters from SBDAE, they are mostly sensitive to words that demonstrate clear polarity. In particular, all the $8$ filters seem to be most activated by certain negative words, and are most deactivated by certain positive words. In this way, the activation of each filter of SBDAE is much more indicative of the polarity than that of DAE, which explains the better performance of SBDAE over DAE. Note that this difference only comes from reweighting the reconstruction errors in a certain way, with no explicit usage of labels. 

\begin{table*}[t]
\caption{Visualization of learned feature maps. From top to bottom: most activated and deactivated words for SBDAE; most activated and deactivated words for DAE.} 
\label{tb:topics}
\scriptsize
\begin{center}
\begin{tabular}{|l|l|l|l|l|l|l|l|}
\hline 
nothing & disappointing & badly & save & even & dull & excuse & ridiculously\\
cannon & worst & disappointing & redeeming & attempt & fails & had & dean\\
outrageously & unfortunately & annoying & awful & unfunny & stupid & failed & none\\
lends & terrible & worst & sucks & couldn't & worst & rest & ruined\\
teacher & predictable & poorly & convince & worst & avoid & he & attempt\\

\hline

first & tears & loved & amazing & excellent & perfect & years & with\\
classic & wonderfully & finest & incredible & surprisingly & ? & terrific & best\\
man & helps & noir & funniest & beauty & powerful & peter & recommended\\
hard & awesome & magnificent & unforgettable & unexpected & excellent & cool & perfect\\
still & terrific & scared & captures & appreciated & favorite & allows & heart\\

\hline  \hline

long & wasn't & probably & to & making & laugh & tv & someone\\
worst & guy & fan & the & give & find & might & yet\\
kids & music & kind & and & performances & where & found & goes\\
anyone & work & years & this & least & before & kids & away\\
trying & now & place & shows & comes & ever & having & poor\\

\hline
done & least & go & kind & recommend & although & ending & worth\\
find & book & trying & takes & instead & everyone & once & interesting\\
before & day & looks & special & wife & anything & wasn't & isn't\\
work & actors & everyone & now & shows & comes & american & rather\\
watching & classic & performances & someone & night & away & sense & around\\

\hline
\end{tabular}
\end{center}

\end{table*}

\section{Related Work and Discussion} \label{related}
Our work falls into the general category of learning representations for text data. In particular, there have been a lot of efforts that try to learn compact representations for either words or documents ~\cite{vsm,lda,lsa,word2vec,pv,mass}. LDA \cite{lda} explicitly learns a set of topics, each of which is defined as a distribution on words; a document is thus represented as the posterior distribution on topics, which is a fixed-length, non-negative vector. Closely related are matrix factorization models such as LSA \cite{lsa} and Non-negative Matrix Factorization (NMF) \cite{nmf}. While LSA factorizes the doc-term matrix via Singular Value Decomposition, NMF learns non-negative basis and coefficient vectors. Similar to these efforts, our model also works directly on the doc-term matrix. However, thanks to the usage of autoencoder, the representation for documents are calculated instantly via direct matrix product, which eliminates the need of expensive inference. Our work also distinguishes itself from other work as a semisupervised representation learning model, where label information can be effectively leveraged.

Recently, there has also been an active thread of research on learning word representations. Notably, \cite{word2vec} shows that we can learn interesting word embeddings via very simple architecture on a large amount of unlabeled dataset. Moreover, \cite{pv} proposed to jointly learn representations for sentences and paragraphs together with words in a similar unsupervised fashion. While our work does not explicitly model the representations for words, it is straightforward to incorporate this idea by adding an additional linear layer at the bottom of the autoencoder.

From the perspective of machine learning methodology, our approach resembles the idea of layer-wise pretraining in deep Neural Networks \cite{bengio}. Our model differs from the traditional training procedure of autoencoders in that we effectively utilize the label information to guide the representation learning. Related idea has been proposed in \cite{semi-ae}, where they train Recursive autoencoders on sentences jointly with prediction of sentiment. Due to the delicate recursive architecture, their model only works on sentences with given parsing trees, and could not generalize to documents. MTC \cite{mtc} is another work that models the interaction of autoencoders and classifiers. However, their training of autoencoders is purely unsupervised, the interaction comes into play by requiring the classifier to be invariant along the tangents of the learned data manifold. It is not difficult to see that the assumption of MTC would not hold when the class labels did not align well with the data manifold, which is a situation our model does not suffer from.

\section{Conclusion} \label{conclusion}
In this paper, we have proposed a novel extension to autoencoders for learning task-specific representations for textual data. We have generalized the traditional autoencoders by relaxing their loss function to the Bregman Divergence, and then derived a discriminative loss function from the label information. Experiments on text classification benchmarks have shown that our model significantly outperforms Bag of Words, traditional Denoising Autoencoder, and other competing methods. We have also qualitatively visualized that our model successfully learns discriminative features, which unsupervised methods fail to do.

\section{Acknowledgments}
This work is supported in part by NSF (CCF-1017828).

\bibliographystyle{aaai}
\bibliography{mybib}{}

\begin{thebibliography}{}

\bibitem[\protect\citeauthoryear{Banerjee \bgroup et al\mbox.\egroup
  }{2004}]{bregman}
Banerjee, A.; Merugu, S.; Dhillon, I.~S.; and Ghosh, J.
\newblock 2004.
\newblock Clustering with bregman divergences.
\newblock In {\em Proceedings of the Fourth {SIAM} International Conference on
  Data Mining, Lake Buena Vista, Florida, USA, April 22-24, 2004},  234--245.

\bibitem[\protect\citeauthoryear{Bengio}{2009}]{bengio}
Bengio, Y.
\newblock 2009.
\newblock Learning deep architectures for {AI}.
\newblock {\em Foundations and Trends in Machine Learning} 2(1):1--127.

\bibitem[\protect\citeauthoryear{Blei, Ng, and Jordan}{2003}]{lda}
Blei, D.~M.; Ng, A.~Y.; and Jordan, M.~I.
\newblock 2003.
\newblock Latent dirichlet allocation.
\newblock {\em Journal of Machine Learning Research} 3:993--1022.

\bibitem[\protect\citeauthoryear{Blitzer, Dredze, and Pereira}{2007}]{amazon}
Blitzer, J.; Dredze, M.; and Pereira, F.
\newblock 2007.
\newblock Biographies, bollywood, boom-boxes and blenders: Domain adaptation
  for sentiment classification.
\newblock In {\em {ACL} 2007, Proceedings of the 45th Annual Meeting of the
  Association for Computational Linguistics, June 23-30, 2007, Prague, Czech
  Republic}.

\bibitem[\protect\citeauthoryear{Deerwester \bgroup et al\mbox.\egroup
  }{1990}]{lsa}
Deerwester, S.~C.; Dumais, S.~T.; Landauer, T.~K.; Furnas, G.~W.; and Harshman,
  R.~A.
\newblock 1990.
\newblock Indexing by latent semantic analysis.
\newblock {\em {JASIS}} 41(6):391--407.

\bibitem[\protect\citeauthoryear{Glorot, Bordes, and Bengio}{2011}]{xavier}
Glorot, X.; Bordes, A.; and Bengio, Y.
\newblock 2011.
\newblock Domain adaptation for large-scale sentiment classification: {A} deep
  learning approach.
\newblock In {\em Proceedings of the 28th International Conference on Machine
  Learning, {ICML} 2011, Bellevue, Washington, USA, June 28 - July 2, 2011},
  513--520.

\bibitem[\protect\citeauthoryear{Le and Mikolov}{2014}]{pv}
Le, Q.~V., and Mikolov, T.
\newblock 2014.
\newblock Distributed representations of sentences and documents.
\newblock In {\em Proceedings of the 31th International Conference on Machine
  Learning, {ICML} 2014, Beijing, China, 21-26 June 2014},  1188--1196.

\bibitem[\protect\citeauthoryear{Maas \bgroup et al\mbox.\egroup }{2011}]{mass}
Maas, A.~L.; Daly, R.~E.; Pham, P.~T.; Huang, D.; Ng, A.~Y.; and Potts, C.
\newblock 2011.
\newblock Learning word vectors for sentiment analysis.
\newblock In {\em The 49th Annual Meeting of the Association for Computational
  Linguistics: Human Language Technologies, Proceedings of the Conference,
  19-24 June, 2011, Portland, Oregon, {USA}},  142--150.

\bibitem[\protect\citeauthoryear{Mikolov \bgroup et al\mbox.\egroup
  }{2013}]{word2vec}
Mikolov, T.; Sutskever, I.; Chen, K.; Corrado, G.~S.; and Dean, J.
\newblock 2013.
\newblock Distributed representations of words and phrases and their
  compositionality.
\newblock In {\em Advances in Neural Information Processing Systems 26: 27th
  Annual Conference on Neural Information Processing Systems 2013. Proceedings
  of a meeting held December 5-8, 2013, Lake Tahoe, Nevada, United States.},
  3111--3119.

\bibitem[\protect\citeauthoryear{Rifai \bgroup et al\mbox.\egroup
  }{2011a}]{mtc}
Rifai, S.; Dauphin, Y.; Vincent, P.; Bengio, Y.; and Muller, X.
\newblock 2011a.
\newblock The manifold tangent classifier.
\newblock In {\em Advances in Neural Information Processing Systems 24: 25th
  Annual Conference on Neural Information Processing Systems 2011. Proceedings
  of a meeting held 12-14 December 2011, Granada, Spain.},  2294--2302.

\bibitem[\protect\citeauthoryear{Rifai \bgroup et al\mbox.\egroup
  }{2011b}]{cae}
Rifai, S.; Vincent, P.; Muller, X.; Glorot, X.; and Bengio, Y.
\newblock 2011b.
\newblock Contractive auto-encoders: Explicit invariance during feature
  extraction.
\newblock In {\em Proceedings of the 28th International Conference on Machine
  Learning, {ICML} 2011, Bellevue, Washington, USA, June 28 - July 2, 2011},
  833--840.

\bibitem[\protect\citeauthoryear{Socher \bgroup et al\mbox.\egroup
  }{2011}]{semi-ae}
Socher, R.; Pennington, J.; Huang, E.~H.; Ng, A.~Y.; and Manning, C.~D.
\newblock 2011.
\newblock Semi-supervised recursive autoencoders for predicting sentiment
  distributions.
\newblock In {\em Proceedings of the 2011 Conference on Empirical Methods in
  Natural Language Processing, {EMNLP} 2011, 27-31 July 2011, John McIntyre
  Conference Centre, Edinburgh, UK, {A} meeting of SIGDAT, a Special Interest
  Group of the {ACL}},  151--161.

\bibitem[\protect\citeauthoryear{Srivastava \bgroup et al\mbox.\egroup
  }{2014}]{dropout}
Srivastava, N.; Hinton, G.~E.; Krizhevsky, A.; Sutskever, I.; and
  Salakhutdinov, R.
\newblock 2014.
\newblock Dropout: a simple way to prevent neural networks from overfitting.
\newblock {\em Journal of Machine Learning Research} 15(1):1929--1958.

\bibitem[\protect\citeauthoryear{Turney and Pantel}{2010}]{vsm}
Turney, P.~D., and Pantel, P.
\newblock 2010.
\newblock From frequency to meaning: Vector space models of semantics.
\newblock {\em J. Artif. Intell. Res. {(JAIR)}} 37:141--188.

\bibitem[\protect\citeauthoryear{Vincent \bgroup et al\mbox.\egroup
  }{2008}]{dae}
Vincent, P.; Larochelle, H.; Bengio, Y.; and Manzagol, P.
\newblock 2008.
\newblock Extracting and composing robust features with denoising autoencoders.
\newblock In {\em Machine Learning, Proceedings of the Twenty-Fifth
  International Conference {(ICML} 2008), Helsinki, Finland, June 5-9, 2008},
  1096--1103.

\bibitem[\protect\citeauthoryear{Wager, Wang, and Liang}{2013}]{lrdrop}
Wager, S.; Wang, S.~I.; and Liang, P.
\newblock 2013.
\newblock Dropout training as adaptive regularization.
\newblock In {\em Advances in Neural Information Processing Systems 26: 27th
  Annual Conference on Neural Information Processing Systems 2013. Proceedings
  of a meeting held December 5-8, 2013, Lake Tahoe, Nevada, United States.},
  351--359.

\bibitem[\protect\citeauthoryear{Xu, Liu, and Gong}{2003}]{nmf}
Xu, W.; Liu, X.; and Gong, Y.
\newblock 2003.
\newblock Document clustering based on non-negative matrix factorization.
\newblock In {\em {SIGIR} 2003: Proceedings of the 26th Annual International
  {ACM} {SIGIR} Conference on Research and Development in Information
  Retrieval, July 28 - August 1, 2003, Toronto, Canada},  267--273.

\end{thebibliography}

\end{document}